\documentclass[11pt]{article}

\usepackage[final]{acl}
\usepackage{times}
\usepackage{latexsym}

\usepackage[T1]{fontenc}

\usepackage[utf8]{inputenc}

\usepackage{microtype}

\usepackage{inconsolata}

\usepackage{graphicx}

\usepackage{amsmath,amssymb}
\usepackage{booktabs}
\usepackage{makecell}
\title{MENTOR: Learning from Failure with On-Policy Pedagogical Intervention}

\author{
  \textbf{Guanlan Yue},
  \textbf{Hang Yu},
  \textbf{Xiaoyong Huang},
  \textbf{ Ruipeng Li}
  \\
  \textbf{Wei Zhang},
  \textbf{Bingchang Liu},
  \textbf{Rui Wang}
  \\[0.5ex]
  \\
  \small{
    \textbf{Correspondence:} \href{mailto:gl2002@stu.xjtu.edu.cn}{gl2002@stu.xjtu.edu.cn}
  }
  }

\begin{document}
\maketitle
\begin{abstract}
Reinforcement Learning from Verifiable Rewards (RLVR) is a key driver for Large Reasoning Models, but it is vulnerable to \textbf{"zero-reward collapse"}, a training failure where a sudden increase in task difficulty causes the learning signal to vanish. We argue that the conventional approach of discarding these failed trajectories is wasteful, as they contain valuable diagnostic information about the model's cognitive gaps, a bright side that can be leveraged for recovery. Existing solutions, however, often regress to static, "off-policy-style" interventions like pre-defined hint libraries or perform post-hoc refinement, failing to provide the adaptive, forward-looking guidance needed to escape the collapse. To address this, we propose a paradigm shift towards \textbf{on-policy pedagogical intervention} and introduce \textbf{MENTOR}, a multi-agent framework that acts as a live mentor. MENTOR diagnoses the root cause of failure in real-time and dynamically generates targeted, instructional prompts to guide the learning agent's next attempt. As a plug-and-play module, MENTOR integrates seamlessly into existing policy optimization pipelines like GRPO without modifying the training infrastructure. Extensive experiments on math and code generation show that MENTOR effectively mitigates zero-reward collapse, outperforming state-of-the-art baselines. Notably, it achieves massive average relative gains of up to 78.3\% on Llama-3.2-3B-Instruct over its base model.
\end{abstract}

\begin{figure}[t]
    \centering
    \includegraphics[page=1, width=0.45\textwidth, keepaspectratio]{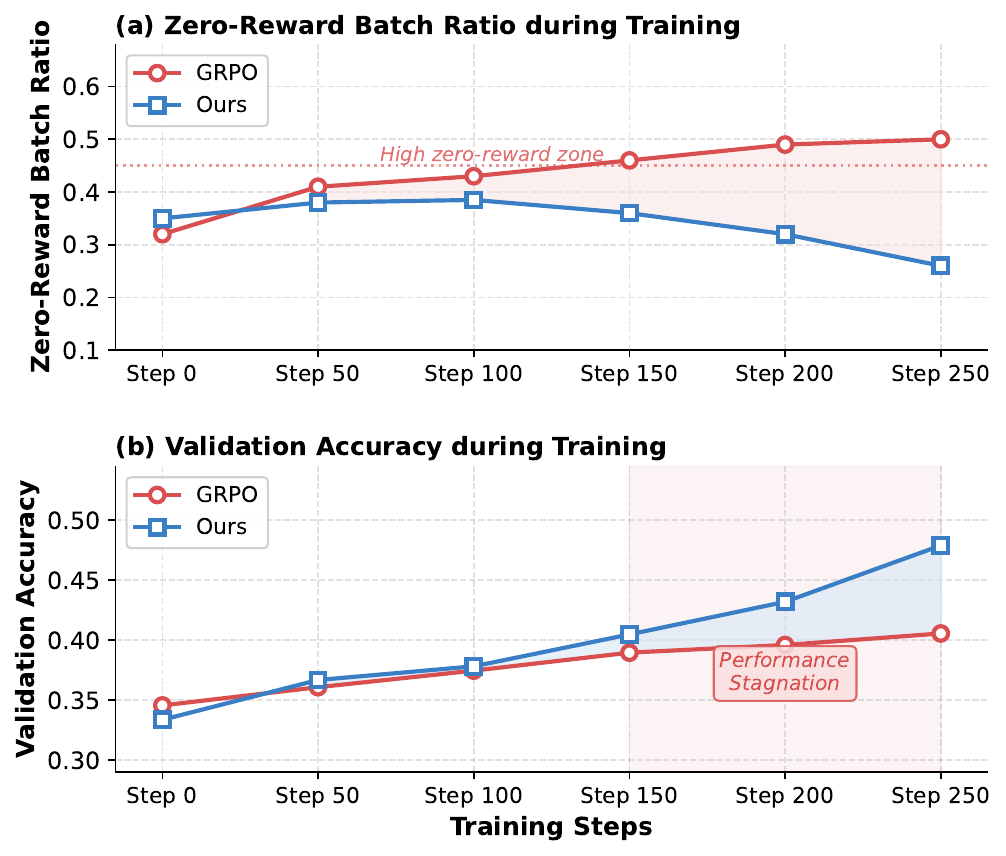}
    \caption{Deepseek-R1-Distill-Qwen-1.5B~\citep{14} math reasoning training dynamics.}
    \label{fig:dynamics}
\end{figure}

\section{Introduction}
Large Reasoning Models have achieved remarkable progress in complex reasoning tasks such as mathematical problem solving and code generation~\citep{1,2,3}. These models generate detailed Chain-of-Thought reasoning traces, exhibit advanced behaviors including self-reflection and self-correction, and demonstrate generalization beyond their training distribution. A key driver of this progress is RLVR, which uses programmatic feedback to encourage exploration of diverse reasoning pathways and iterative strategy refinement.

However, RLVR training is vulnerable to a critical failure mode. When task difficulty exceeds the model's current capability boundary, all sampled reasoning trajectories can fail simultaneously. This results in zero reward for the entire batch, causing the learning signal to vanish and halting policy improvement. This phenomenon, which we term ``zero-reward collapse'', arises from the misalignment between task complexity and model competence~\citep{4,5}. As illustrated in Figure~\ref{fig:dynamics}, once collapse occurs, the fraction of zero-reward samples rises rapidly and model performance plateaus immediately.

Yet failed reasoning trajectories are not without value: each contains diagnostic information about where the error occurred, what cognitive deficiency it reflects, and what correction is needed. Discarding them entirely prevents the system from capturing real-time error patterns, leading to teaching signals misaligned with the true root causes of failure.
This challenge exposes a fundamental tension in modern RL alignment. The field has progressed from off-policy methods like DPO~\citep{dpo}, which learn from static datasets, to \textbf{on-policy methods like GRPO}~\citep{9}, which learn adaptively from real-time exploration. While on-policy learning is superior, its reliance on occasional success for gradient signals becomes its Achilles' heel during zero-reward collapse. As compared in Figure~\ref{fig:teaser}, existing solutions for this collapse fall short in several ways. Some resort to \textbf{simply discarding the failed batches entirely} (e.g., DAPO~\citep{6}), which wastes valuable diagnostic signals. Others regress to an \textbf{``off-policy'' style of intervention}: they use \textbf{static, pre-generated hint libraries} (e.g., LUFFY~\citep{luffy}, QUESTA~\citep{7}, HINT~\citep{25}) or \textbf{pre-defined curricula} (Scaf-GRPO~\citep{16}). Even more advanced methods that perform \textbf{post-hoc refinement} of failed trajectories focus on output correction and data augmentation rather than providing the forward-looking guidance needed to escape the collapse (e.g., Critique-GRPO~\citep{critique}, SRPO~\citep{srpo}). These static and reactive approaches cannot adapt to the model's specific, evolving failure modes, creating a ``distributional drift'' where the guidance becomes misaligned and ineffective. We argue that \textbf{the solution is not to abandon on-policy learning, but to augment it. What is needed is on-policy pedagogical intervention}, a live mentor that diagnoses the root cause of failure in real-time and provides a targeted teaching signal to help the policy escape the collapse.

To realize this vision, we propose MENTOR (\textbf{M}ulti-agent \textbf{E}rror-aware \textbf{N}avigation and \textbf{T}eaching for \textbf{O}nline \textbf{R}einforcement learning), a framework that operationalizes on-policy pedagogical intervention. It enhances RLVR by converting moments of systemic failure into targeted teaching opportunities. An Attribution Agent identifies conceptual gaps and logical errors in failed trajectories; a Generation Agent creates targeted instructional prompts based on this diagnosis; a Judgment Agent validates prompt quality via formal rules and LLM-as-a-Judge evaluation; and a Repair Agent iteratively refines prompts using feedback. Together, they form a closed-loop teaching mechanism that integrates seamlessly into policy optimization pipelines such as GRPO without modifying the training infrastructure. Our main contributions are as follows:

\begin{itemize}
   \item We identify and formally characterize zero-reward collapse, a fundamental training bottleneck in RLVR. We argue that overcoming it requires a paradigm shift from static, off-policy-style fixes to dynamic, on-policy pedagogical intervention, exposing a shared structural limitation in existing RLVR methods.
   \item We propose MENTOR, an end-to-end multi-agent framework for \textbf{online pedagogical intervention} in RLVR. To our knowledge, it is the first to embed feedback-driven, context-aware instructional prompting directly into the RLVR training loop. As a plug-and-play module, MENTOR constructs, validates, and iteratively refines targeted prompts without pre-built corpora, ensuring supervision remains aligned with the model's current cognitive state throughout training.
   \item We evaluate MENTOR on mathematical reasoning and code generation. Results show that MENTOR consistently outperforms state-of-the-art baselines, achieving massive average relative gains of up to 78.3\% on Llama-3.2-3B-Instruct over its base model. MENTOR \textbf{effectively mitigates training stagnation from zero-reward collapse and restores meaningful gradient signals}.
\end{itemize}

\section{Related Work}

\begin{figure}[t]
    \centering
    \includegraphics[width=0.45\textwidth, keepaspectratio]{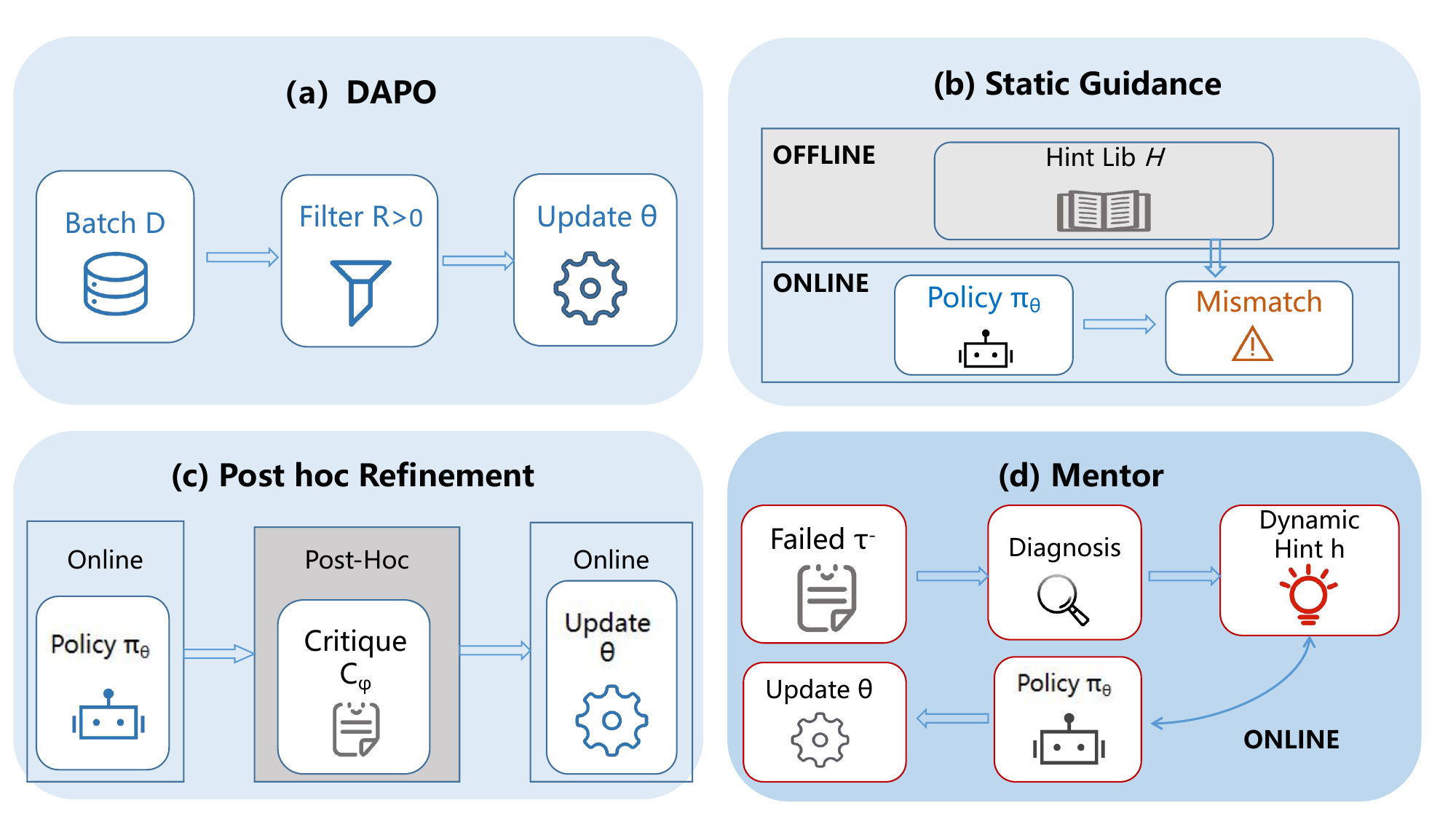}
    \caption{Comparison of mitigation strategies for zero-reward collapse in RLVR.}
    \label{fig:teaser}
\end{figure}

The problem of zero-reward collapse has spurred various mitigation strategies. However, as illustrated in Figure~\ref{fig:teaser},   existing approaches can be categorized by their underlying intervention philosophy, which, when analyzed, reveals a shared structural gap. They fall short of providing the \textbf{on-policy pedagogical intervention} necessary to resolve systemic training failure.

\subsection{Failure Avoidance via Filtering} 
The most direct strategy is to simply sidestep the problem. Methods like \textbf{DAPO}~\citep{6} employ hard-example filtering, identifying and discarding training batches where all rewards are zero. While this prevents the model from receiving degenerate gradient signals, it is a fundamentally passive approach. It treats systemic failure as an anomaly to be ignored rather than a symptom to be diagnosed. By filtering out the most challenging examples, this strategy offers no mechanism for the model to learn from its deepest cognitive errors and push its capability boundary.

\subsection{Static ``Off-Policy'' Guidance}
A more proactive set of methods attempts to provide external guidance, but does so using static, pre-generated resources, analogous to off-policy learning in RLHF. These approaches range from injecting high-quality reasoning traces from an offline corpus, as seen in \textbf{LUFFY}~\citep{luffy} and \textbf{QUESTA}~\citep{7}, or pre-written conceptual heuristics as in \textbf{HINT}~\citep{25}, to employing pre-defined curricula. For instance, \textbf{Scaf-GRPO}~\citep{16} uses a pre-built, hierarchical library of hints that are selected based on the model's performance plateaus. While more structured than simple filtering, all these methods are fundamentally limited by their static nature. They rely on a fixed corpus of guidance that cannot adapt to the model's specific, evolving failure modes during on-policy training. As the policy changes, its behavior inevitably diverges from the distribution of the offline data, leading to \textbf{``distributional drift''}. The pre-canned guidance becomes progressively less relevant, losing its effectiveness and potentially even destabilizing the learning process.

\subsection{Post-Hoc Refinement and Critique}
A more advanced paradigm focuses on learning from failure by refining or critiquing the model's own outputs. This category includes a family of methods such as \textbf{Critique-GRPO~\citep{critique}, SRPO~\citep{srpo}, CTRL~\citep{ctrl}, and Critique-RL~\citep{citquerl}}. These frameworks analyze a failed trajectory and either generate a critique to refine it for data augmentation \textbf{(Critique-GRPO)}, train the policy to perform internal self-correction \textbf{(SRPO)}, or train a general-purpose critic model \textbf{(CTRL, Critique-RL)}. Although these methods are dynamic, their focus is \textbf{retrospective and corrective, not prospective and pedagogical}. They aim to ``fix the answer'' after the fact to create a better training sample. They do not, however, generate a targeted, forward-looking instructional hint designed to guide the model's next reasoning attempt on the same problem. This is a critical distinction: they improve the data, but they do not directly teach the model the underlying cognitive skill it is missing, making them less effective at resolving systemic collapse.

\subsection{Distinctions from Foundational Methodologies}
Our work also builds upon and distinguishes itself from several foundational methodologies. For instance, the use of an \textbf{LLM-as-a-Judge}~\citep{20} for automated evaluation and feedback-based prompt refinement at the inference stage~\citep{21} are powerful tools. However, MENTOR is the first to integrate this capability into a \textbf{closed-loop, online RL training pipeline} specifically to resolve training collapse. Similarly, our approach shares conceptual ground with \textbf{knowledge distillation}, where experience is transferred from a teacher to a student model~\citep{opd1, opd3}. Yet, these methods typically rely on static, offline distillation and lack the ability to diagnose and fix inherent cognitive deficiencies in real-time~\citep{opd4, opd5}. In contrast, MENTOR's framework can be seen as a highly dynamic, \textbf{failure-driven form of distillation}, generating teaching signals only when and where they are needed.

\subsection{The Unaddressed Gap: The Need for On-Policy Pedagogical Intervention}
Collectively, these strategies highlight a crucial, unaddressed gap in RLVR. \textbf{Failure avoidance} is too passive. \textbf{Static guidance} suffers from distributional drift. \textbf{Post-hoc refinement} is corrective, not pedagogical. None of these paradigms provide what is truly needed when an on-policy agent gets stuck: \textbf{real-time, diagnostic intervention}. No existing method performs a cognitive-level diagnosis of why the policy is systemically failing and then dynamically generates a targeted, instructional hint to guide its subsequent exploration. This is the core innovation of MENTOR. By establishing a closed-loop, multi-agent system for \textbf{on-policy pedagogical intervention}, MENTOR moves beyond avoidance, static guidance, and post-hoc refinement to actively teach the model how to overcome its current limitations, directly addressing the root cause of zero-reward collapse.

\section{Method}

Our method, MENTOR, is designed to overcome the "zero-reward collapse" problem in RLVR frameworks like GRPO. As illustrated in Figure~\ref{fig2}, MENTOR operates as a plug-and-play module that intercepts moments of systemic failure and transforms them into pedagogical opportunities. Instead of relying on static hint libraries, MENTOR generates instructional prompts on-demand, dynamically tailoring the teaching signal to the policy's evolving cognitive gaps. The process consists of three main stages: (1) real-time detection of collapsed samples, (2) dynamic generation of pedagogical hints via a multi-agent system, and (3) seamless integration of these hints back into the training loop. Details of prompts are shown in Appendix~\ref{sec:appendixa}.

\subsection{Collapse-Aware Sample Filtering (CASF)}
\label{sec:casf}

The first stage is to identify instances of zero-reward collapse. This occurs when for a given problem instance $q_i$, all $N$ trajectories $\{\tau_j^{(i)}\}_{j=1}^N$ sampled from the policy $\pi_\theta$ receive a reward of zero. Such a scenario leads to degenerate advantages and a vanishing gradient signal, halting learning for that instance. For example, consider the math problem $q_i$: "Determine whether there exists a nowhere dense set in $\mathbb{R}$ that is not an $F_\sigma$ set." As detailed in Table~\ref{tab:mentor_case_study} in Appendix~\ref{appendixb}, the model might generate $N=4$ trajectories, all incorrectly concluding "No" based on various flawed arguments.

Since all trajectories fail verification, they all receive a reward of zero, triggering a collapse for this instance. To detect this, we introduce a lightweight hook into the training loop that executes after the reward scoring step. For each batch of training data, the hook identifies all instances where every sampled trajectory failed and collects them into a "collapse set" $C$. If $C$ is empty, training proceeds normally. Otherwise, for each collapsed instance $(q_i, a_i) \in C$ (where $a_i$ is the ground-truth answer), we preserve the full set of failed trajectories $\{\tau_j^{(i)}\}_{j=1}^N$. These trajectories, while yielding no reward, contain valuable diagnostic information, revealing recurring error patterns that point to the model's underlying misconceptions. The resulting triplet $(q_i, a_i, \{\tau_j^{(i)}\}_{j=1}^N)$ for each collapsed instance is then passed to our guidance generation system.

\subsection{Failure-Driven Contextual Guidance (FDCG)}
\label{sec:failure_driven_guidance}

To transform the diagnostic information from failed trajectories into a high-quality learning signal, we introduce a multi-agent system for FDCG. This system collaboratively synthesizes and validates targeted instructional hints. Continuing our example from Table~\ref{tab:mentor_case_study} in Appendix~\ref{appendixb}, the FDCG system receives the problem, the ground-truth answer ("Yes"), and the set of four failed trajectories. Formally, for each collapsed instance, the FDCG system, denoted by $\mathcal{A}$, generates a validated hint $h_i^*$ by processing the failed trajectories in the context of the current policy $\pi_\theta$:
\begin{equation}
    h_i^* = \mathcal{A}(q_i, a_i, \{\tau_j^{(i)}\}_{j=1}^N; \pi_\theta)
\end{equation}
This process is carried out by a sequence of specialized agents:

\begin{figure}[t]
    \centering
    \includegraphics[page=1, width=0.48\textwidth, keepaspectratio]{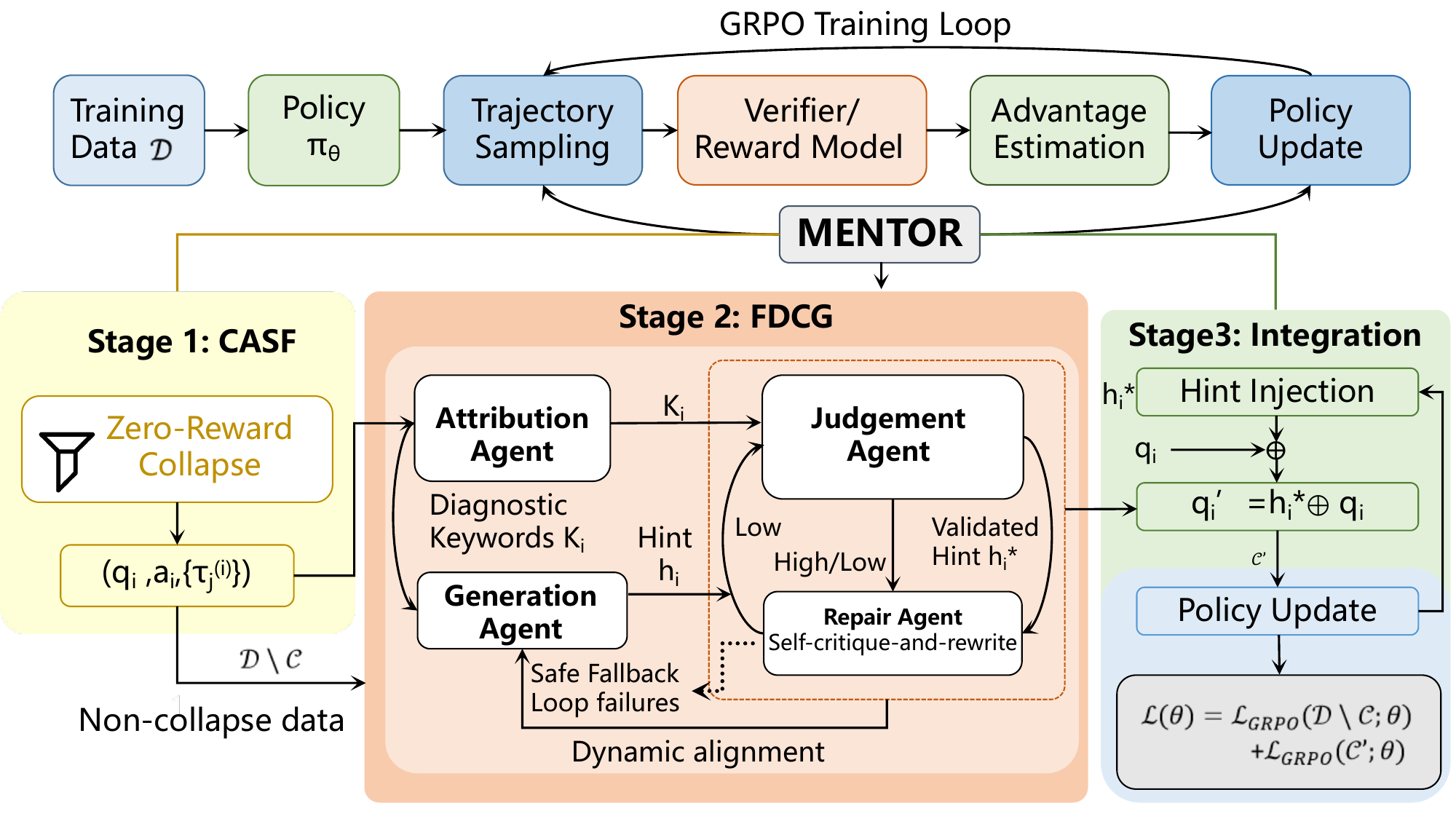}
    \caption{Overview of the MENTOR framework.}
    \label{fig2}
\end{figure}
\noindent\textbf{Attribution Agent.} To provide targeted, rather than generic, guidance, we must first diagnose the specific cognitive error. The Attribution Agent is responsible for this diagnosis. Given the problem $q_i$, answer $a_i$, and the set of failed trajectories $\{\tau_j^{(i)}\}_{j=1}^N$, it analyzes the reasoning chains to identify persistent failure causes. In our example (Table~\ref{tab:mentor_case_study}), the agent analyzes the failed attempts and identifies that the model consistently misunderstands the relationship between topological properties, outputting a set of diagnostic keywords $K_i$: "Confused nowhere dense sets with $F_\sigma$ ...". These keywords localize the conceptual gap without revealing the solution path. As the policy $\pi_\theta$ evolves, so do its failure modes, ensuring that the diagnosis from $K_i$ remains tightly aligned with the model's current deficiencies.

\noindent\textbf{Generation Agent.} Based on the diagnosis, the Generation Agent acts as a pedagogical tutor. Using the diagnostic keywords $K_i$ as a directive and the full set of failed trajectories as context, it generates a candidate instructional hint $\hat{h}_i$. For instance, using the keywords from our example, the agent generates a candidate hint $\hat{h}_i$: "Consider whether every nowhere dense ...". It focuses on correcting the cognitive error while deliberately avoiding answer leakage or excessive reasoning steps, thereby encouraging the policy to rediscover the solution path itself.

\noindent\textbf{Judgment-Repair Loop.} To ensure that generated hints are helpful, safe, and pedagogically sound, they undergo a rigorous closed-loop validation and refinement process. As shown in Table~\ref{tab:mentor_case_study} in Appendix~\ref{appendixb}, this initial hint undergoes a two-round refinement. In Round 1, the Judgment Agent evaluates the hint as "LOW" quality, critiquing it as "too vague". The Repair Agent uses this critique to generate a more directive version: "Hint:  Think about whether a nowhere ...". In Round 2, the Judgment Agent validates this refined hint as "HIGH" quality, making it the final output $h_i^*$. To prevent invalid teaching signals from entering the training pipeline, if a hint repeatedly fails validation, it is replaced by a pre-defined safe fallback hint. This two-tiered mechanism guarantees the pedagogical efficacy and safety of the final validated hint $h_i^*$.

\subsection{Integration into GRPO Training}
\label{sec:grpo_integration}

The final stage seamlessly injects the validated pedagogical guidance back into the GRPO training process to recover the learning signal.

 \noindent\textbf{Hint Injection.} For each collapsed sample $(q_i, a_i) \in C$, we form an augmented input by prepending the validated hint $h_i^*$ to the original problem query: $q'_i = h_i^* \oplus q_i$, where $\oplus$ denotes concatenation. In our example, the new augmented input $q'_i$ becomes: "Hint: Think about whether a nowhere... Original Question: Determine whether there..."

 \noindent\textbf{Policy Update.} This enhanced query $q'_i$ replaces the original $q_i$ and re-enters the standard GRPO training pipeline. With this targeted, pedagogical nudge, the policy $\pi_\theta$ is now significantly more likely to generate trajectories that receive non-zero rewards. As shown by the "New Rollout" section in Table~\ref{tab:mentor_case_study}, the guided policy now generates several successful trajectories (e.g., New Rollouts 1, 2, and 4), correctly answering "Yes" by constructing a non-Borel subset of the Cantor set. These successful attempts receive non-zero rewards, restoring a meaningful gradient for this previously unsolvable problem. Crucially, MENTOR is training-objective-agnostic. It modifies only the input data for collapsed samples, leaving the GRPO loss function, optimizers, and reward mechanism entirely unchanged. The total loss is simply computed over the original non-collapsed data $D \setminus C$ and the new augmented collapsed data $C'$:
 \begin{equation}
    \mathcal{L}(\theta) = \mathcal{L}_{\text{GRPO}}(D \setminus C; \theta) + \mathcal{L}_{\text{GRPO}}(C'; \theta)
\end{equation}
where $C' = \{(q'_i, a_i) | (q_i, a_i) \in C\}$. This plug-and-play nature allows MENTOR to be integrated into any GRPO variant without altering the core training infrastructure, fundamentally mitigating zero-reward collapse by delivering targeted, interpretable, and dynamically aligned teaching signals. Detailed case studies are presented in Appendix~\ref{appendixb}.
\begin{table*}[t]
\centering
\setlength{\tabcolsep}{1mm}
\small
\renewcommand{\arraystretch}{0.4} 
\begin{tabular}{lccccccc|c}
\toprule
Methods & AIME24 & AIME25 &  MATH-500 & GAOKAO-Bench  & Olympiad-Bench  & DM & Avg & MMLU \\
\midrule
\cmidrule{1-9}
\multicolumn{9}{c}{\textbf{Deepseek-R1-Distill-Qwen-1.5B}} \\
Base          & 16.67          & 16.67          & 66.20          & 63.55          & 2.50           & 34.32          & 38.57          & 17.14          \\
GRPO  & 16.67          & 16.67          & 67.20          & 65.89          & 2.00           & 40.88          & 43.55          & 21.43          \\
DAPO          & 13.33          & 16.67          & 71.20          & 72.90          & 3.50           & 43.72          & 46.66          & 22.86          \\
SRPO          & \underline{20.00} & 16.67          & 71.40          & 72.90          & 3.50           & 43.84          & 46.83          & 17.14          \\
Critique & 16.67          & 16.67          & 71.60          & \underline{74.70} & 2.50           & 43.88          & 46.92          & 24.29          \\
QUESTA        & \underline{20.00} & \underline{20.00} & 67.20          & 72.42          & 3.00           & 43.76          & 46.14          & 24.29          \\
LUFFY         & 16.67          & 16.67          & 67.60          & 72.90          & 3.00           & 43.80          & 46.20          & 15.71          \\
HINT          & \underline{20.00} & 16.67          & 71.00          & 73.36          & 3.50           & \underline{44.20} & 47.06          & 24.29          \\
Scaf-GRPO     & \underline{20.00} & 16.67          & \underline{72.40} & 73.80          & \underline{4.00} & 44.00          & \underline{47.18} & \underline{25.71} \\
Ours          & \textbf{30.00} & \textbf{23.33} & \textbf{75.80} & \textbf{78.50} & \textbf{4.50} & \textbf{49.60} & \textbf{52.16} & \textbf{28.57} \\
              & {(+80.0\%)} &{(+40.0\%)} & {(+14.5\%)} & {(+23.5\%)} & {(+80.0\%)} & {(+44.5\%)} & {(+35.2\%)} & {(+66.7\%)} \\
\cmidrule{1-9}
\multicolumn{9}{c}{\textbf{Llama-3.2-3B-Instruct}} \\
Base          & 3.33           & 3.33           & 32.20          & 0.93           & 1.50           & 19.08          & 18.57          & 5.71           \\
GRPO  & 10.00          & \underline{6.67} & \underline{43.20} & 10.75          & 2.50           & 27.92          & 27.26          & 14.29          \\
DAPO          & 10.00          & \underline{6.67} & \underline{43.20} & \underline{12.15} & 2.50           & 29.86          & 28.74          & \underline{15.71} \\
SRPO          & 16.67          & 3.33           & 40.00          & 7.94           & 2.00           & 30.04          & 28.15          & 11.43          \\
Critique & 10.00          & \underline{6.67} & 41.20          & 7.94           & \underline{3.00} & 30.20          & 28.47          & 14.29          \\
QUESTA        & 16.67          & \underline{6.67} & 40.65          & 7.95           & 2.50           & 30.06          & 28.32          & 12.85          \\
LUFFY         & 13.33          & \underline{6.67} & 40.20          & 10.28          & 2.00           & 30.00          & 28.30          & 14.29          \\
HINT          & 13.33          & \underline{6.67} & 41.59          & 10.28          & 2.50           & 30.40          & 28.81          & \underline{15.71} \\
Scaf-GRPO     & \underline{20.00} & \underline{6.67} & 42.52          & 7.85           & 2.50           & \underline{30.50} & \underline{28.93} & \underline{15.71} \\
Ours          & \textbf{26.67} & \textbf{13.33} & \textbf{44.60} & \textbf{16.50} & \textbf{5.50} & \textbf{34.76} & \textbf{33.11} & \textbf{18.57} \\
              & {(+700.9\%)} & {(+300.3\%)} & {(+38.5\%)} & {(+1674.2\%)} & {(+266.7\%)} &{(+82.2\%)} & {(+78.3\%)} & {(+225.2\%)} \\
\cmidrule{1-9}
\multicolumn{9}{c}{\textbf{Qwen3-4B}} \\
Base          & 60.00          & 53.33          & 87.80          & 79.91          & 28.50          & 82.64          & 79.65          & 68.57          \\
GRPO  & 60.00          & 56.67          & \underline{88.60} & \underline{90.65} & 28.50          & 85.84          & 82.76          & 71.43          \\
DAPO          & \underline{66.67} & 50.00          & 88.40          & 90.46          & \underline{31.50} & 86.92          & 83.67          & 71.43          \\
SRPO          & \underline{66.67} & 56.67          & 86.40          & 87.85          & 30.00          & 87.04          & 83.28          & 70.00          \\
Critique & 63.33          & \underline{60.00} & 86.60          & 90.18          & 30.69          & 87.08          & 83.52          & 72.86          \\
QUESTA        & 63.33          & \underline{60.00} & 87.20          & 83.18          & 29.50          & 86.76          & 82.87          & 72.86          \\
LUFFY         & 56.67          & 53.33          & 86.80          & 81.78          & 29.00          & 86.92          & 82.70          & 68.57          \\
HINT          & 60.00          & 56.67          & 87.40          & 89.71          & 29.50          & 87.24          & 83.59          & \underline{74.29} \\
Scaf-GRPO     & \underline{66.67} & \underline{60.00} & 88.20          & 82.24          & 30.84          & \underline{87.80} & \underline{83.81} & \underline{74.29} \\
Ours          & \textbf{73.33} & \textbf{70.00} & \textbf{88.80} & \textbf{91.12} & \textbf{32.50} & \textbf{92.00} & \textbf{87.71} & \textbf{78.57} \\
              & {(+22.2\%)} & {(+31.3\%)} & {(+1.1\%)} & {(+14.0\%)} & {(+14.0\%)} & {(+11.3\%)} & {(+10.1\%)} & {(+14.6\%)} \\
\cmidrule{1-9}
\multicolumn{9}{c}{\textbf{Qwen2.5-Math-7B}} \\
Base          & 20.00          & 6.67           & 65.60          & 1.87           & \underline{6.00} & 37.88          & 37.39          & 10.00          \\
GRPO  & 13.33          & 10.00          & 71.00          & \underline{8.69} & 4.00           & 52.16          & 48.72          & 14.28          \\
DAPO          & 20.00          & 10.00          & 72.40          & 8.41           & 4.00           & 53.16          & 49.68          & \underline{18.57}          \\
SRPO          & 20.00          & 10.00          & 72.40          & 8.41           & 4.00           & 52.36          & 49.11          & 17.14          \\
Critique & \underline{23.33} & 10.00          & 72.80          & 6.90           & 4.50           & 54.20          & 50.45          & \textbf{20.00} \\
QUESTA        & 20.00          & 10.00          & 71.80          & 6.90           & 4.50           & 53.80          & 49.99          & 15.71          \\
LUFFY         & 20.00          & 6.67           & 70.60          & 7.01           & 3.50           & 53.60          & 49.60          & 12.86          \\
HINT          & \underline{23.33} & \underline{13.33} & \underline{74.80} & 7.10           & 5.50           & 54.20          & 50.84          & 15.71          \\
Scaf-GRPO     & \underline{23.33} & 10.00          & 72.80          & 6.90           & 5.00           & \underline{54.92} & \underline{51.00} & \underline{18.57}          \\
Ours          & \textbf{30.00} & \textbf{20.00} & \textbf{76.00} & \textbf{12.15} & \textbf{7.50} & \textbf{60.50} & \textbf{56.09} & \textbf{20.00} \\
              & {(+50.0\%)} & {(+199.9\%)} & {(+15.9\%)} & {(+549.7\%)} & {(+25.0\%)} & {(+59.7\%)} & {(+50.0\%)} & {(+100.0\%)} \\
\cmidrule{1-9}
\end{tabular}%
\caption{Performance (\%) on math reasoning benchmarks and out-of-distribution (OOD) generalization on MMLU. Best: bold; second-best: underlined.Avg denotes the sample-weighted average.All improvements reported are relative gains.}
\label{tab:combined_compact}
\end{table*}

\begin{table}[t]
\centering
\setlength{\tabcolsep}{0.2pt}
\small
\renewcommand{\arraystretch}{0.3} 
\begin{tabular}{llllll}
\toprule
Methods & LCB & HE & MBPP & BCB & Avg \\
\cmidrule{1-6}
\multicolumn{6}{c}{\textbf{Deepseek-Coder-1.3B}} \\
Base            & 61.93          & 15.85          & 46.40          & 20.09          & 30.13          \\
GRPO   & 64.77          & 16.46          & 49.40          & 23.60          & 33.18         \\
DAPO            & 65.34          & 17.07          & 50.00          & 23.86          & 33.59          \\
SRPO            & 65.34          & 18.29          & 49.60          & 24.04          & 33.69          \\
Critique   & 65.34          & 18.90          & 50.00          & \underline{24.56} & 34.14          \\
QUESTA          & 65.91          & 19.51          & 49.60          & 24.30          & 33.99          \\
LUFFY           & 66.48          & 19.51         & 49.60          & 24.21          & 33.99          \\
HINT            & 66.48          & 19.51          & 50.40          & 24.30          & \underline{34.24} \\
Scaf-GRPO       & \underline{67.05} & \underline{20.12} & \underline{51.00} & 23.86          & \underline{34.24} \\
Ours            & \textbf{69.89} & \textbf{24.39} & \textbf{54.60} & \textbf{27.02} & \textbf{37.58} \\
                & {(+12.9\%)} & {(+53.9\%)} & {(+17.7\%)} & {(+34.5\%)} & {(+24.7\%)} \\
\cmidrule{1-6}
\multicolumn{6}{c}{\textbf{Deepseek-R1-Distill-Qwen-1.5B}} \\
Base            & 32.95          & 15.85          & 24.60          & 11.67          & 17.17          \\
GRPO  & 35.80          & 18.29          & 27.60          & 15.18          & 20.40          \\
DAPO            & 36.36          & 19.51          & 28.20          & 15.44          & 20.86          \\
SRPO            & 36.36          & 18.90          & 28.40          & 15.53          & 20.91          \\
Critique  & 36.36          & 18.90          & 28.40          & 15.61          & 20.96          \\
QUESTA          & 36.93          & 19.51          & 28.20          & 15.44          & 20.91          \\
LUFFY           & 37.50          & 18.90          & 28.00          & 15.44          & 20.86          \\
HINT            & 37.50          & 19.51          & 28.60          & \underline{16.14} & 21.46          \\
Scaf-GRPO       & \underline{38.07} & \underline{20.12} & \underline{29.20} & 15.88          & \underline{21.57} \\
Ours            & \textbf{40.91} & \textbf{24.39} & \textbf{32.80} & \textbf{18.60} & \textbf{24.65} \\
                & {(+24.2\%)} &{(+53.9\%)} & {(+33.3\%)} &{(+59.4\%)} & {(+43.6\%)} \\
\cmidrule{1-6}
\multicolumn{6}{c}{\textbf{Llama-3.2-3B-Instruct}} \\
Base            & 62.50          & 37.20          & 46.40          & 12.81          & 27.71          \\
GRPO    & 65.34          & 39.63          & 49.40          & 16.32          & 30.96          \\
DAPO            & 65.91          & 40.85          & 50.00          & 16.58          & 31.41          \\
SRPO            & 65.91          & 40.24          & 50.20          & 16.67          & 31.46          \\
Critique & 65.91          & 40.85          & 50.20          & 16.75          & 31.56          \\
QUESTA          & 67.05          & 40.24          & 50.00          & 16.58          & 31.46          \\
LUFFY           & 66.48          & 40.24          & 49.80          & 16.58          & 31.36          \\
HINT            & 67.05          & \underline{41.46} & 50.40          & 17.02          & 31.92          \\
Scaf-GRPO       & \underline{67.61} & 40.85          & \underline{51.00} & \underline{17.28} & \underline{32.22} \\
Ours            & \textbf{70.45} & \textbf{45.73} & \textbf{54.60} & \textbf{19.74} & \textbf{35.20} \\
                & {(+12.7\%)} & {(+22.9\%)} & {(+17.7\%)} & {(+54.1\%)} & {(+27.0\%)} \\
\cmidrule{1-6}
\multicolumn{6}{c}{\textbf{Qwen3-4B}} \\
Base            & 35.23          & 20.12          & 35.60          & 13.42          & 21.52          \\
GRPO    & 38.07          & 22.56          & 38.60          & 16.93          & 24.75          \\
DAPO            & 38.64          & 23.78          & 39.20          & 17.19          & 25.20          \\
SRPO            & 39.20          & 23.17          & 39.40          & 17.28          & 25.30          \\
Critique  & \underline{40.34} & 23.17          & 39.40          & \underline{17.89} & 25.76          \\
QUESTA          & 39.77          & 23.78          & 39.20          & 17.28          & 25.35          \\
LUFFY           & 39.20          & 23.17          & 39.00          & 17.19          & 25.15          \\
HINT            & 39.77          & 23.78          & \underline{40.20} & 17.63          & \underline{25.81} \\
Scaf-GRPO       & 38.64          & \underline{24.39} & 39.60          & \underline{17.89} & 25.76          \\
Ours            & \textbf{43.18} & \textbf{28.66} & \textbf{43.80} & \textbf{20.35} & \textbf{28.99} \\
                & {(+22.6\%)} & {(+42.4\%)} & {(+23.0\%)} & {(+51.6\%)} & {(+34.7\%)} \\
\bottomrule
\end{tabular}

\caption{Performance (\%) on code generation benchmarks. Best: bold; second-best: underlined.Avg denotes the sample-weighted average.All improvements reported are relative gains.}
\label{tab:coderesult}
\end{table}

\section{Experiments}
\subsection{Experimental Setup}

 \noindent\textbf{Tasks and Models.} We train MENTOR on complex reasoning in \textbf{mathematical problem-solving} (DeepMath-103K~\citep{22}, 10K train/2.5K valid) and \textbf{code generation} (LiveCodeBench~\citep{23}, 1,055 instances, 5:1 train/valid split). We test on a diverse set of models to assess cross-architecture transferability: \textbf{math-specialized} (Qwen2.5-Math-7B~\citep{qwen2}), \textbf{code-specialized} (DeepSeek-Coder-1.3B~\citep{DeepSeek-Coder}), \textbf{lightweight} (DeepSeek-R1-Distill-Qwen-1.5B~\citep{14}), and \textbf{general-purpose} (Qwen3-4B~\citep{qwen3}, Llama-3.2-3B-Instruct~\citep{llama}). For evaluation, we use a suite of standard benchmarks, including AIME~\citep{aime24,aime25}, MATH-500~\citep{math500}, GAOKAO-Bench~\citep{gaokao} and Olympiad-Bench~\citep{olmpiadbench} for math, and HumanEval~\citep{humaneval}(HE), MBPP~\citep{MBPP}, and BigCodeBench~\citep{bigcodebench}(BCB) for code. We also test out-of-distribution generalization on MMLU~\citep{mmlu}.

\noindent\textbf{Training Details.} All experiments are conducted on eight H20-3E GPUs. We train for one epoch with a learning rate of $1 \times 10^{-6}$ and sample four trajectories per problem. \textbf{The pedagogical agents within MENTOR (e.g., for hint generation and critique) are powered by Qwen3-32B.}

 \noindent\textbf{Baselines.} We compare MENTOR against eight State-of-the-Art(SOTA) baselines representing the primary paradigms for mitigating zero-reward collapse. These include the standard on-policy algorithm with \textbf{no mitigation}, (1) Vanilla GRPO(GRPO) \citep{9}; \textbf{a failure avoidance} method, (2) DAPO \citep{6}, which filters out zero-reward batches; \textbf{post-hoc refinement} methods such as (3) Critique-GRPO(Critique) \citep{critique} and (4) SRPO \citep{srpo}, which critique or refine failed trajectories for data augmentation; and finally, methods using \textbf{static, off-policy guidance} from pre-defined hint libraries or curricula, including (5) LUFFY \citep{luffy}, (6) QUESTA \citep{7}, (7) HINT \citep{25}, and (8) Scaf-GRPO \citep{16}.
\textbf{The hints required by all baselines are generated by the same Qwen3-32B model under a matched generation budget}.

\subsection{Main Results}
As shown in Tables~\ref{tab:combined_compact} and~\ref{tab:coderesult}, MENTOR consistently outperforms all baseline strategies across both mathematical reasoning and code generation tasks, validating the superiority of on-policy pedagogical intervention. 

Compared to Vanilla GRPO, which often stagnates due to zero-reward collapse, MENTOR achieves massive average relative gains, such as \textbf{+19.8\% on DeepSeek-R1-Distill-Qwen-1.5B for math and +20.8\% for code}. This demonstrates that actively converting failures into teaching moments is far more effective than simply waiting for sporadic successes. MENTOR also surpasses passive \textbf{failure avoidance} strategies like DAPO, confirming our core hypothesis that \textbf{discarding failed trajectories is wasteful, and leveraging their diagnostic information for targeted intervention is critical for recovery}.

Furthermore, MENTOR consistently outperforms methods using static, \textbf{off-policy guidance} like QUESTA, HINT, and Scaf-GRPO. For instance, on the math-specialized Qwen2.5-Math-7B, MENTOR achieves a \textbf{50.0\%} relative gain, surpassing all static baselines. This confirms that \textbf{dynamic, on-policy adaptation to the model's evolving failure modes is superior to static guidance that suffers from distributional drift}. Similarly, MENTOR proves more effective than post-hoc refinement methods like Critique-GRPO. Its forward-looking, pedagogical guidance teaches the model the underlying skill it is missing, proving more effective at resolving systemic collapse than the retrospective, corrective feedback of refinement-based approaches.
MENTOR's impact is particularly pronounced on smaller or less-specialized models, which are more susceptible to zero-reward collapse. On the lightweight DeepSeek-R1-Distill-Qwen-1.5B, MENTOR boosts math accuracy by \textbf{35.2\%} and code accuracy by \textbf{43.6\%} relative to the base model. On the general-purpose Llama-3.2-3B-Instruct, it achieves a remarkable \textbf{78.3\%} relative gain in math. These results show that \textbf{MENTOR's targeted instructional signals effectively overcome training stagnation and unlock the latent reasoning potential of these models}. Even on highly specialized models like DeepSeek-Coder-1.3B, MENTOR delivers a \textbf{24.7\%} relative gain, showing that its failure-attribution-driven teaching is beneficial even for tuned models.
To assess whether MENTOR learns robust reasoning rather than in-domain pattern matching, we evaluate out-of-distribution (OOD) generalization on MMLU. As shown in Table~\ref{tab:combined_compact}, MENTOR consistently outperforms or matches the best baselines on MMLU across all four models (e.g., achieving a \textbf{66.7\%} relative gain on DeepSeek-R1-Distill-Qwen-1.5B). This strong transfer to unseen, knowledge-intensive tasks indicates that \textbf{MENTOR helps the model capture deeper, generalizable reasoning structures rather than superficial correlations, significantly enhancing model robustness}.

\subsection{Efficiency and Cost-Benefit Analysis}
\label{sec:time_cost}
MENTOR offers a practical and scalable solution by automating the pedagogical process. As illustrated in Table~\ref{tab:time_cost}, unlike strong hint-based baselines such as Scaf-GRPO and QUESTA, which rely on costly and labor-intensive offline data preparation (about 24 expert hours), MENTOR eliminates this manual effort entirely. The reported costs include Qwen3-32B multi-agent inference. Moreover, compared to other online methods like Critique-GRPO that suffer from computational bottlenecks, \textbf{MENTOR is approximately 24\% faster in wall-clock time (17.9h vs. 23.5h)}. While the FDCG module introduces a marginal computational overhead compared to vanilla GRPO (a modest 2.6 hours on Qwen3-4B), this small investment represents a highly effective trade-off. 
This is because it (1) \textbf{Delivers significant performance improvements}, achieving average relative gains of up to +78.3\% (Llama-3.2-3B-Instruct on math), and (2) \textbf{Enables dynamic adaptivity} by producing high-quality, targeted hints to prevent reward collapse and adapt to the model's evolving failure modes, a crucial capability that vanilla GRPO lacks entirely.
\begin{table}[ht]
\centering
\small
\setlength{\tabcolsep}{3pt}
\begin{tabular}{lcccc}
\toprule
\textbf{Method} & \begin{tabular}{@{}c@{}}\textbf{GPU} \\ \textbf{Hours}\end{tabular} & \begin{tabular}{@{}c@{}}\textbf{Wall} \\ \textbf{Time}\end{tabular} & \begin{tabular}{@{}c@{}}\textbf{Human} \\ \textbf{Labor}\end{tabular} & \begin{tabular}{@{}c@{}}\textbf{Total} \\ \textbf{Time}\end{tabular} \\
\midrule
Vanilla GRPO & 120h & 15.3h & 0h & 15.3h \\
Critique-GRPO & 168h & 23.5h & 0h & 23.5h \\
Offline Baselines & 132h & 21.3h & $\sim$24h\textsuperscript{*} & $\sim$45.3h \\
MENTOR (Ours) & 134.7h & 17.9h & \textbf{0h (Auto)} & \textbf{17.9h} \\
\bottomrule
\multicolumn{5}{l}{\textsuperscript{*}\small Estimated expert hours for data curation.} \\
\end{tabular}
\caption{Training time and labor cost comparison.} 
\label{tab:time_cost}
\end{table}

\subsection{Ablation and Hyperparameter Analysis}
\subsubsection{Ablation of Core Agents.}
We conduct an ablation study on Qwen3-4B for math reasoning (Table~\ref{tab:ablation}) to validate the contribution of each agent in MENTOR's collaborative framework.

\begin{table}[ht]
\centering
\small
\setlength{\tabcolsep}{3pt}
\begin{tabular}{lcc}
\toprule
\textbf{Method} & \textbf{Avg. Acc. (\%)} & \textbf{$\delta$Acc. (\%)} \\
\midrule
w/o Attribution Agent        & 83.21 & -4.50 \\
w/o Generation Agent         & 83.04 & -4.67 \\
w/o Critique-Repair Agent & 85.43 & -2.28 \\
Ours                         & 87.71 & - \\
\bottomrule
\end{tabular}
\caption{Ablation study on core components.}
\label{tab:ablation}
\end{table}

 \noindent\textbf{w/o Attribution Agent:} Removing the diagnosis agent causes a \textbf{4.50\%} drop in accuracy. Without precise error diagnosis, the Generation Agent receives noisy, unfocused context, \textbf{weakening the efficacy of the teaching signal}.

 \noindent\textbf{w/o Generation Agent:} Directly using the diagnostic keywords from the Attribution Agent as a hint yields the worst performance, with a \textbf{4.67\%} drop. This shows that \textbf{diagnostic insights must be translated into structured, pedagogical prompts to be effective}.

 \noindent\textbf{w/o Judgment-Repair Loop:} Bypassing the validation and refinement loop leads to a \textbf{2.28\%} decrease in accuracy. This confirms the necessity of the loop's quality control, as \textbf{unverified prompts can be misleading or suboptimal, undermining the learning process}.
\subsubsection{Hyperparameter Analysis.}
\begin{figure}[t]
    \centering
    \includegraphics[width=0.48\textwidth, keepaspectratio]{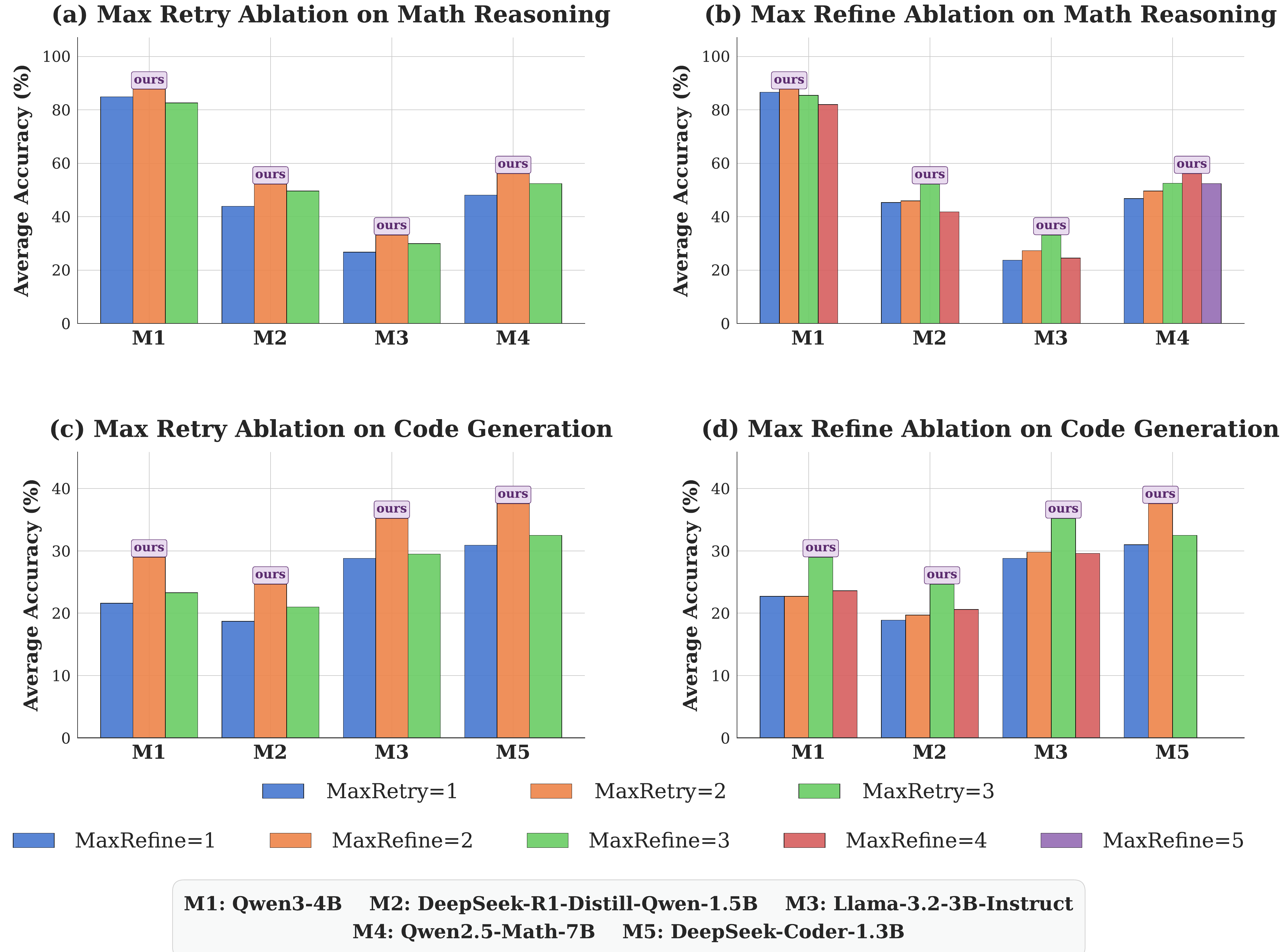}
    \caption{Hyperparameter ablation of Max Retry and Refine.}
    \label{fig:hyperparam}
\end{figure}

We analyze two key hyperparameters, Max Retry (number of full hint generation attempts) and Max Refine (internal refinement iterations), as shown in Figure~\ref{fig:hyperparam}.

 \noindent\textbf{Max Retry.}
Across both math and code, performance consistently peaks at Max Retry = 2. A value of 1 lacks fault tolerance against poor initial hints, while a value of 3 introduces redundant or noisy prompts. This shows that a \textbf{balanced number of attempts is crucial for generating diverse yet high-quality guidance}.

\noindent\textbf{Max Refine.}
The optimal Max Refine value varies by model and task, reflecting differences in error complexity. For instance, Qwen3-4B peaks at Max Refine = 2 for math, while the smaller Llama-3.2-3B-Instruct benefits from more refinement (Max Refine = 3). This indicates that \textbf{more complex or subtle error patterns require deeper semantic analysis and restructuring within the Judgment-Repair loop}.
\section{Conclusion}
We propose MENTOR to address the training bottleneck in RLVR caused by zero-reward collapse. It dynamically identifies policy failures during online learning, attributes underlying cognitive errors, and guides the agent with structured teaching prompts for targeted policy refinement, thereby surpassing the learning limits of existing approaches. Experiments show that MENTOR achieves strong generalization across model scales and significantly outperforms state-of-the-art baselines on challenging benchmarks in mathematical reasoning and code generation. By actively focusing on hard examples and enabling learning from failure, MENTOR not only enhances complex task performance but also overcomes the training ceiling imposed by zero-reward collapse.

\section*{Limitations}
The current prompt generation and reasoning guidance mechanism is only validated in single-modal scenarios, while its applicability in complex dynamic environments and multi-modal fusion tasks has not been fully explored.

\bibliography{custom}

\appendix

\section{Prompt}
\label{sec:appendixa}
Table~\ref{tab:unified_keyphrase} to Table~\ref{tab:rewrite_prompt} present the prompt templates adopted in our experiments.
These prompts implement the Attribution, Generation, Judgment, and Repair agents in FDCG: they diagnose failures and produce/validate instructional \emph{hints}.

\begin{table*}[h!]
\centering
\small
\begin{tabular}{p{\dimexpr\linewidth-2\tabcolsep\relax}}
\hline
\textbf{ERROR\_KEYPHRASE\_PROMPT} (Attribution Agent) \\
\hline
You are an Attribution Agent. Given a question, its ground-truth answer, and one or more failed student solutions/trajectories, diagnose the persistent cognitive error. \\
Output diagnostic keywords that localize the conceptual gap or logical breakpoint. Do NOT reveal the solution path, final answer, formulas, or implementation steps. \\
Focus on logical and algorithmic mistakes rather than trivial syntax issues. \\
Prefer a short keyword phrase (or a semicolon-separated set of keywords) capturing the root misconception. \\
Examples: \\
- Confused nowhere dense sets with $F_\sigma$ sets \\
- Off-by-one error in loop condition \\
- Incorrect base case in recursion \\
- Misapplied mathematical formula \\
- Wrong algorithm selection for task \\
Question: \{question\} \\
Ground truth: \{answer\} \\
Failed Trajectories: \{model\_output\} \\
Diagnostic Keywords: \\
\hline
\end{tabular}
\caption{Prompt template for Attribution Agent (diagnostic keyword extraction)}
\label{tab:unified_keyphrase}
\end{table*}

\begin{table*}[h!]
\centering

\small
\begin{tabular}{p{\dimexpr\linewidth-2\tabcolsep\relax}}
\hline
\textbf{GUIDANCE\_PROMPT\_TEMPLATE} (Generation Agent) \\
\hline
You are a Generation Agent acting as a pedagogical tutor. Using the diagnostic keywords as a directive and the failed trajectories as context, write ONE candidate instructional hint for the same question. \\
The hint should: \\
1. Directly address the diagnosed misconception or logical breakpoint \\
2. Be concise, actionable, and specific to the failure mode \\
3. Guide thinking without giving the solution path \\
4. Avoid answer leakage: no final answers, numerical results, formula derivations, or concrete code snippets \\
5. Avoid excessive intermediate reasoning steps; encourage the learner to rediscover the solution \\
Question: \{question\} \\
Diagnostic Keywords: \{error\_keyphrase\} \\
Failed Trajectories: \{model\_output\} \\
Candidate Hint: \\
\hline
\end{tabular}
\caption{Prompt template for Generation Agent (instructional hint generation)}
\label{tab:unified_guidance}
\end{table*}

\begin{table*}[h!]
\centering

\small
\begin{tabular}{p{\dimexpr\linewidth-2\tabcolsep\relax}}
\hline
\textbf{HINT\_VERIFY\_PROMPT} (Judgment Agent) \\
\hline
You are a Judgment Agent (LLM-as-a-Judge). Evaluate whether a \emph{generated hint} is HIGH or LOW quality. Do NOT revise the student's solution; judge only the hint. \\
A HIGH quality hint: \\
- Semantically aligned with the diagnosed failure (diagnostic keywords) and the question \\
- Points to the core concept, principle, algorithm, or reasoning pattern that failed \\
- Does not leak the answer: no final answer, formula derivation, calculation, or complete implementation \\
- Clear and targeted, not vague or generic \\
A LOW quality hint: \\
- Misaligned with the diagnosed failure, irrelevant, ambiguous, or misleading \\
- Reveals answers, formulas, detailed solving steps, or solution code \\
- Overly generic without actionable pedagogical guidance \\
Respond with EXACTLY two lines and no extra content: \\
Quality: HIGH \\
Reason: N/A \\
Or for low-quality hints: \\
Quality: LOW \\
Reason: One sentence explaining the specific defect. \\
Question: \{question\} \\
Diagnostic Keywords: \{error\_keyphrase\} \\
Failed Trajectories: \{model\_output\} \\
Hint: \{hint\} \\
Quality: \\
\hline
\end{tabular}
\caption{Prompt template for Judgment Agent (hint quality verification)}
\label{tab:unified_verify}
\end{table*}

\begin{table*}[h!]
\centering

\small
\begin{tabular}{p{\dimexpr\linewidth-2\tabcolsep\relax}}

\hline
\textbf{HINT\_REFINE\_PROMPT} (Repair Agent) \\
\hline
You are a Repair Agent. Revise the \emph{hint} (not the student's solution) according to the Judgment critique so it becomes a safe, pedagogically sound instructional prompt. \\
Question: \{question\} \\
Diagnostic Keywords: \{error\_keyphrase\} \\
Original Hint: \{current\_hint\} \\
Critique / Issue(s): \{fixed\_feedback\} \\
Guidelines: \\
- Fully resolve all listed defects \\
- Keep semantic alignment with the diagnosed failure \\
- Restrict output to one or two clear and complete sentences \\
- Do not include concrete formulas, derivation steps, calculations, code snippets, or the final answer \\
Output strictly following the format: \\
Your refined reasoning. \\
Hint: Your refined hint. \\
Now generate your response following the exact format above: \\
\hline
\end{tabular}
\caption{Prompt template for Repair Agent (hint refinement)}
\label{tab:unified_refine}
\end{table*}

\begin{table*}[h!]
\centering

\small
\begin{tabular}{p{\dimexpr\linewidth-2\tabcolsep\relax}}

\hline
\textbf{CRITIQUE\_PROMPT} (Repair Agent: Critique step) \\
\hline
You are a Repair Agent performing self-critique on an instructional hint. Identify EXACTLY ONE critical flaw that prevents this hint from being HIGH quality. \\
Choose from: \\
- Too vague (e.g., no specific concept named) \\
- Reveals the answer or key number (answer leakage) \\
- Gives step-by-step solution instructions \\
- Misaligned with the diagnosed failure / mentions irrelevant concept \\
- Is longer than two sentences \\
- Uses formulas or markdown that leak the solution \\
Respond in one sentence starting with ``Flaw: ". \\
Question: \{question\} \\
Diagnostic Keywords: \{error\_keyphrase\} \\
Hint: \{current\_hint\} \\
\hline
\end{tabular}
\caption{Prompt template for Repair Agent critique step}
\label{tab:critique_prompt}
\end{table*}

\begin{table*}[h!]
\centering

\small
\begin{tabular}{p{\dimexpr\linewidth-2\tabcolsep\relax}}

\hline
\textbf{REWRITE\_PROMPT} (Repair Agent: Rewrite step) \\
\hline
Revise the following \emph{hint} to fix the flaw: ``\{flaw\}". Do not revise the student's solution. \\
Question: \{question\} \\
Diagnostic Keywords: \{error\_keyphrase\} \\
Original Hint: \{current\_hint\} \\
Guidelines: \\
- Fix the stated flaw \\
- Keep alignment with the diagnosed failure \\
- Keep to 1--2 clear, complete sentences \\
- Do NOT include formulas, steps, calculations, or the answer \\
- If the flaw is ``Too vague'', name the specific theorem/principle/concept without revealing the solution \\
Output EXACTLY: \\
Your reasoning. \\
Hint: Your revised hint. \\
\hline
\end{tabular}
\caption{Prompt template for Repair Agent rewrite step}
\label{tab:rewrite_prompt}
\end{table*}

\section{Case}
\label{appendixb}
Table~\ref{tab:mentor_case_study} to Table~\ref{tab:mentor_case_study_geometry} present three instantiations of the MENTOR framework.

\begin{table*}[ht]
\centering
\small
\setlength{\tabcolsep}{0.9pt}
\begin{tabular}{@{}p{\linewidth}@{}}
\toprule
\textbf{Input:} ``Determine whether there exists a nowhere dense set in $\mathbb{R}$ that is not an $F_\sigma$ set.'' \\
\midrule
\textbf{Rollout} \\
\textit{Rollout 1:} ``A nowhere dense set has empty interior in its closure... Therefore all nowhere dense sets are $F_\sigma$. Answer: No.'' \\
\textit{Rollout 2:} ``$F_\sigma$ sets are countable unions of closed sets... and nowhere dense sets are meager, the answer is No.'' \\
\textit{Rollout 3:} ``The Cantor set is closed and nowhere dense... Therefore all nowhere dense sets are $F_\sigma$. Answer: No.'' \\
\textit{Rollout 4:} ``By the Baire Category Theorem, any meager... set is contained in a countable union of closed sets... Answer: No.''\\
\midrule
\textbf{Attribution Agent:} ``Confused nowhere dense sets with $F_\sigma$ sets; misapplied Baire Category Theorem by equating meager sets with $F_\sigma$ sets.'' \\
\midrule
\textbf{Generation Agent:} ``Consider whether every nowhere dense set can be expressed as a countable union of closed sets...'' \\
\midrule
\textbf{Judge Agent(Round 1):} ``Quality: LOW \newline Reason: too vague; does not guide toward a specific counterexample...'' \\
\textbf{Repair Agent(Critique):} ``Flaw: Too vague --- does not name a specific concept (e.g., sets beyond the Borel hierarchy)...'' \\
\textbf{Repair Agent(Rewrite):} ``Hint: Think about whether a nowhere dense set constructed beyond the Borel hierarchy could fail to be an $F_\sigma$.'' \\
\textbf{Judge Agent(Round 2):} ``Quality: HIGH\newline  Reason: N/A'' \\
\midrule
\textbf{Hint Injection \& New Rollout} \\
\textit{New Rollout 1:} ``The hint points to subsets beyond the Borel hierarchy. Let $C$ be the Cantor set, which is closed and nowhere dense. Any $A \subseteq C$ has $\overline{A} \subseteq C$, so $A$ is nowhere dense; since $C$ has cardinality continuum it admits a non-Borel subset $A$, which cannot be $F_\sigma$. Answer: Yes.''\\
\textit{New Rollout 2:} ``Every $F_\sigma$ set is Borel. The Cantor set $C$ is nowhere dense with $2^{\aleph_0}$ points, so it has $2^{2^{\aleph_0}}$ subsets but only $2^{\aleph_0}$ Borel ones; hence some $A \subseteq C$ is non-Borel, thus not $F_\sigma$, yet nowhere dense. Answer: Yes.'' \\
\textit{New Rollout 3:}  ``While the hint mentions the Borel hierarchy, I cannot construct a definitive non-$F_\sigma$ nowhere dense set... Answer: No.''  \\
\textit{New Rollout 4:}  ``Take a non-Borel subset $A$ of the Cantor set $C$, whose existence follows from a cardinality argument. As a subset of a nowhere dense set $A$ is nowhere dense, but being non-Borel it is not $F_\sigma$. Answer: Yes.''  \\
\bottomrule
\end{tabular}
\caption{MENTOR framework: complete process on the question ``Determine whether there ... is not an $F_\sigma$ set'' and ground-truth answer ``Yes", with 4 rollouts.}
\label{tab:mentor_case_study}
\end{table*}

\begin{table*}[h!]
\centering
\small
\setlength{\tabcolsep}{10pt}
\renewcommand{\arraystretch}{1.4}
\begin{tabular}{p{14cm}}
\toprule

\textbf{Input:} ``Find the number of ordered pairs $(x,y)$, where both $x$ and $y$ are integers between $-100$ and $100$, inclusive, such that $12x^{2}-xy-6y^{2}=0$.'' \\
\midrule 
\textbf{Rollout} \\
\textit{Rollout 1:}
``Factoring the equation $12x^2-xy-6y^2 = 0$, we get $(3x+2y)(4x-3y)=0$. Case 1: $3x+2y=0 \implies y = -1.5x$. Letting $x=2k, y=-3k$, $-100 \le -3k \le 100 \implies -33 \le k \le 33$, which gives 67 pairs. Case 2: $4x-3y=0 \implies y = \frac{4}{3}x$. Letting $x=3m, y=4m$, $-100 \le 4m \le 100 \implies -25 \le m \le 25$, which gives 51 pairs. Total number of pairs is $67 + 51 = 118$. Answer: 118.'' \\

\textit{Rollout 2:}
``We have $12x^2-xy-6y^2 = 0$. Using the quadratic formula for $x$, $x = \frac{y \pm \sqrt{y^2 - 4(12)(-6y^2)}}{24} = \frac{y \pm 17y}{24}$. So $x = \frac{3y}{4}$ or $x = \frac{-2y}{3}$. For $x = 3y/4$, $y$ is a multiple of 4, giving 51 solutions. For $x = -2y/3$, $y$ is a multiple of 3, giving 67 solutions. The total is $51 + 67 = 118$. Answer: 118.'' \\

\textit{Rollout 3:}
``The quadratic factors to $(3x + 2y)(4x - 3y) = 0$. There are 67 integer solutions for the first line within the range $[-100, 100]$ and 51 integer solutions for the second line. Total pairs = 118. Answer: 118.'' \\

\textit{Rollout 4:}
``$(3x+2y)(4x-3y)=0$. From $3x=-2y$, there are 67 solutions. From $4x=3y$, there are 51 solutions. Adding them gives $118$ solutions. I don't see any other constraints. Answer: 118.'' \\

\midrule

\textbf{Attribution Agent} \\
\textit{Output $K_i$:} ``Overlapping solution sets treated as disjoint; shared solutions counted twice when the two linear cases are added.'' \\

\midrule

\textbf{Generation Agent} \\
\textit{Output $\hat{h}_i$:} ``Think about whether any ordered pair is counted in both linear cases and needs to be subtracted from the total sum.'' \\

\midrule

\textbf{Judge Agent(Round 1)} 
\textit{Output:} ``Quality: LOW\newline Reason: The hint is too generic and does not explicitly name the intersection point (the origin) to guide the student to the exact double-counted case.'' \\

\textbf{Repair Agent(Critique)} 
\textit{Output:} ``Flaw: Too vague --- the hint does not explicitly name the exact overlapping point that might be counted twice in both linear relationships.'' \\

\textbf{Repair Agent(Rewrite)}
\textit{Output:} ``Naming the origin turns the vague warning about overlap into a concrete check, without prescribing the arithmetic correction or revealing the final count. Hint: You counted the pairs on each line separately, so examine whether a point such as the origin $(0,0)$ satisfies both linear equations before you combine the two counts.'' \\

\textbf{Judge Agent(Round 2)}
\textit{Output:} ``Quality: HIGH\newline Reason: N/A'' \\

\midrule

\textbf{Hint Injection} \\
\textit{Augmented input $q'_i = h_i^* \oplus q_i$, where $h_i^*$ is the hint accepted in Round 2:} ``Hint: You counted the pairs on each line separately, so examine whether a point such as the origin $(0,0)$ satisfies both linear equations before you combine the two counts. Original Question: Find the number of ordered pairs $(x,y)$, where both $x$ and $y$ are integers between $-100$ and $100$, inclusive, such that $12x^{2}-xy-6y^{2}=0$.'' \\

\midrule

\textbf{New Rollout} \\
\textit{New Rollout 1:} ``Factoring gives $(3x+2y)(4x-3y)=0$. Case 1 gives 67 pairs and Case 2 gives 51 pairs. Checking the origin as the hint suggests, $(0,0)$ satisfies both equations, so it has been counted twice and must be removed once. Total = $67 + 51 - 1 = 117$. Answer: 117.''\\
 
\textit{New Rollout 2:} ``$x = 3y/4$ has 51 solutions, and $x = -2y/3$ has 67 solutions. Following the hint, $(0,0)$ is counted twice. The total unique solutions are $51 + 67 - 1 = 117$. Answer: 117.'' \\

\textit{New Rollout 3:}  ``The two lines $y = -1.5x$ and $y = 4x/3$ intersect at the origin $(0,0)$. We count 67 points on the first line and 51 points on the second line. Because the origin is on both, $67 + 51 - 1 = 117$. Answer: 117.''  \\
 
\textit{New Rollout 4:}  ``From $3x+2y=0$ we get 67 points, and from $4x-3y=0$ we get 51 points. Testing the point named in the hint, $(0,0)$ satisfies both equations, so it is double counted and subtracting 1 gives $117$. Answer: 117.''  \\

\bottomrule
\end{tabular}
\caption{MENTOR framework: complete step-by-step process on the question ``Find the number of ordered pairs $(x,y)$, where both $x$ and $y$ are integers between $-100$ and $100$, inclusive, such that $12x^{2}-xy-6y^{2}=0$'' and ground-truth answer ``117", with 4 rollouts.}
\label{tab:mentor_case_study_math}
\end{table*}

\begin{table*}[h!]
\centering
\small
\setlength{\tabcolsep}{10pt}
\renewcommand{\arraystretch}{1.4}
\begin{tabular}{p{14cm}}
\toprule

\textbf{Input:} ``An isosceles trapezoid has an inscribed circle tangent to each of its four sides. The radius of the circle is 3, and the area of the trapezoid is 72. Let the parallel sides of the trapezoid have lengths $r$ and $s$, with $r \neq s$. Find $r^{2}+s^{2}$.'' \\
\midrule
\textbf{Rollout} \\
\textit{Rollout 1:}
``The inradius gives the height $h=6$, and the area gives $\frac{1}{2}(r+s)(6) = 72 \implies r+s=24$. Because a circle is inscribed, the two legs satisfy $2c = r+s$, so each leg is $c=12$. A leg rises by the height and runs horizontally by the difference of the two bases, so $(s-r) = \sqrt{12^2-6^2} = \sqrt{108}$. Therefore $r^2+s^2 = \frac{(r+s)^2+(s-r)^2}{2} = \frac{576+108}{2} = 342$. Answer: 342.'' \\

\textit{Rollout 2:}
``Place the longer base from $(0,0)$ to $(s,0)$ and the shorter one from $(s-r,6)$ to $(s,6)$, so the slanted leg joins $(0,0)$ to $(s-r,6)$. With $r+s=24$ and each leg equal to 12 by the tangency condition, $(s-r)^2 + 6^2 = 12^2 \implies (s-r)^2 = 108$, giving $r^2+s^2 = \frac{576+108}{2} = 342$. Answer: 342.'' \\

\textit{Rollout 3:}
``Height $h=6$, $r+s=24$. By Pitot's theorem for tangential quadrilaterals, the slant side $c$ satisfies $2c = r+s = 24$, so $c=12$. Drawing altitudes forms a right triangle with hypotenuse 12, height 6, and base $s-r$. By the Pythagorean theorem, $(s-r)^2 + 6^2 = 12^2 \implies (s-r)^2 = 108$. We know $2(r^2+s^2) = (r+s)^2 + (s-r)^2 = 576 + 108 = 684$. So $r^2+s^2 = 342$. Answer: 342.'' \\

\textit{Rollout 4:}
``$r+s=24$ and $c=12$. The right triangle from the altitude has legs $6$ and $s-r$, and hypotenuse $12$. So $(s-r)^2 = 12^2 - 6^2 = 108$. We want $r^2+s^2$. Since $r+s=24$ and $(s-r)^2=108$, we add them: $2r^2+2s^2 = 576 + 108 = 684 \implies r^2+s^2 = 342$. Answer: 342.'' \\

\midrule

\textbf{Attribution Agent} \\
\textit{Output $K_i$:} ``Misidentified the horizontal leg of the altitude-based right triangle; ignored the symmetry of the two congruent outer segments of the longer base.'' \\

\midrule

\textbf{Generation Agent} \\
\textit{Output $\hat{h}_i$:} ``Remember that when you drop altitudes from the top base to the bottom base of an isosceles trapezoid, the base of the resulting right triangle is $(s-r)/2$, not $s-r$.'' \\

\midrule

\textbf{Judge Agent(Round 1)} 
\textit{Output:} ``Quality: LOW\newline Reason: The hint directly points out the exact formula correction, which borders on step-by-step solution instructions rather than guiding the student to discover the geometric relationship themselves.'' \\

\textbf{Repair Agent(Critique)} 
\textit{Output:} ``Flaw: Gives step-by-step solution instructions by explicitly providing the algebraic expression $(s-r)/2$ for the triangle's base.'' \\

\textbf{Repair Agent(Rewrite)}
\textit{Output:} ``Asking the student to visualize or draw the altitudes and explicitly identify the lengths of the segments on the bottom base encourages them to find the correct triangle base themselves. Hint: Draw the altitudes from the top vertices to the bottom base and carefully determine the length of the base of the resulting right triangles in terms of $r$ and $s$.'' \\

\textbf{Judge Agent(Round 2)}
\textit{Output:} ``Quality: HIGH\newline Reason: N/A'' \\

\midrule

\textbf{Hint Injection} \\
\textit{Augmented input $q'_i = h_i^* \oplus q_i$, where $h_i^*$ is the hint accepted in Round 2:} ``Hint: Draw the altitudes from the top vertices to the bottom base and carefully determine the length of the base of the resulting right triangles in terms of $r$ and $s$. Original Question: An isosceles trapezoid has an inscribed circle tangent to each of its four sides. The radius of the circle is 3, and the area of the trapezoid is 72. Let the parallel sides of the trapezoid have lengths $r$ and $s$, with $r \neq s$. Find $r^{2}+s^{2}$.'' \\

\midrule

\textbf{New Rollout} \\
\textit{New Rollout 1:} ``Area is 72, $h=6$, so $r+s=24$. By Pitot's theorem, $r+s=2c \implies c=12$. Following the hint, dropping the altitudes splits the bottom base $s$ into three segments: a central one of length $r$, and two equal outer segments. Each right triangle's base is $(s-r)/2$. Thus, $((s-r)/2)^2 + 6^2 = 12^2 \implies ((s-r)/2)^2 = 108 \implies (s-r)^2 = 432$. We know $2(r^2+s^2) = (r+s)^2 + (s-r)^2 = 576 + 432 = 1008 \implies r^2+s^2 = 504$. Answer: 504.''\\
 
\textit{New Rollout 2:} ``$r+s=24$ and the slant side is $c=12$. Drawing the altitudes as the hint suggests reveals the right triangle has a base of $(s-r)/2$. By the Pythagorean theorem, $((s-r)/2)^2 = 12^2 - 6^2 = 108$, giving $(s-r)^2 = 432$. Then $r^2+s^2 = \frac{(r+s)^2 + (s-r)^2}{2} = \frac{576 + 432}{2} = 504$. Answer: 504.'' \\

\textit{New Rollout 3:}  ``$h=6 \implies r+s=24$. The tangential property gives the side length 12. Following the hint, I drop the altitudes and see that the outer segments are equal, so the horizontal leg is $(s-r)/2$ rather than $s-r$. So $((s-r)/2)^2 = 108 \implies (s-r)^2 = 432$. Finally, $r^2+s^2 = ((24)^2 + 432)/2 = 504$. Answer: 504.''  \\
 
\textit{New Rollout 4:}  ``$r+s=24$, slant side $c=12$. Drawing the altitudes as the hint asks, the longer base splits into a central segment of length $r$ and two equal outer ones, so each right triangle has base $(s-r)/2$. Therefore $((s-r)/2)^2 + 36 = 144 \implies (s-r)^2 = 432$. The value $r^2+s^2$ equals half the sum of $(r+s)^2$ and $(s-r)^2$, which is $(576+432)/2 = 504$. Answer: 504.''  \\

\bottomrule
\end{tabular}
\caption{MENTOR framework: complete step-by-step process on the question ``An isosceles trapezoid has an inscribed circle tangent to each of its four sides. The radius of the circle is 3, and the area of the trapezoid is 72. Let the parallel sides of the trapezoid have lengths $r$ and $s$, with $r \neq s$. Find $r^{2}+s^{2}$'' and ground-truth answer ``504", with 4 rollouts.}
\label{tab:mentor_case_study_geometry}
\end{table*}

\end{document}